\documentclass[letterpaper, 10 pt, conference]{ieeeconf}

\IEEEoverridecommandlockouts                              

\usepackage{graphics} 
\usepackage{amsmath} 
\usepackage{amssymb}  
\usepackage{subcaption}
\usepackage{graphicx}
\usepackage{float}

\usepackage{algorithm}
\usepackage[noend]{algpseudocode}

\algrenewcommand\algorithmicforall{\textbf{for each}}
\algrenewcommand\algorithmicindent{.8em}

\usepackage[pdftex,dvipsnames]{xcolor}  

\algdef{SE}[DOWHILE]{Do}{doWhile}{\algorithmicdo}[1]{\algorithmicwhile\ #1}%
\algnewcommand\algorithmicforeach{\textbf{for each}}
\algdef{S}[FOR]{ForEach}[1]{\algorithmicforeach\ #1\ \algorithmicdo}

\usepackage[bookmarks=false]{hyperref}

\hypersetup{
colorlinks=true,
linkcolor=blue,
filecolor=magenta,
urlcolor=black,
}

\title{\LARGE \bf
Material Mapping in Unknown Environments using Tapping Sound
}

\author{Shyam Sundar Kannan$^{1}$, Wonse Jo$^{1}$, Ramviyas Parasuraman$^{2}$, and Byung-Cheol Min$^{1}$
\thanks{$^{1}$Shyam Sundar Kannan, Wonse Jo, and Byung-Cheol Min are with SMART Lab, Department of Computer and Information Technology, Purdue University, West Lafayette, IN 47907, USA \tt\small{kannan9@purdue.edu, jow@purdue.edu, minb@purdue.edu}}%
\thanks{$^{2}$Ramviyas Parasuraman is with the Department of Computer Science, University of Georgia, Athens, GA 30602, USA {\tt\small ramviyas@uga.edu}}%
}

\begin{document}

\maketitle
\thispagestyle{empty}
\pagestyle{empty}

\begin{abstract}
In this paper, we propose an autonomous exploration and a tapping mechanism-based material mapping system for a mobile robot in unknown environments. The goal of the proposed system is to integrate simultaneous localization and mapping (SLAM) modules and sound-based material classification to enable a mobile robot to explore an unknown environment autonomously and at the same time identify the various objects and materials in the environment. This creates a material map that localizes the various materials in the environment which has potential applications for search and rescue scenarios. A tapping mechanism and tapping audio signal processing based on machine learning techniques are exploited for a robot to identify the objects and materials. We demonstrate the proposed system through experiments using a mobile robot platform installed with Velodyne LiDAR, a linear solenoid, and microphones in an exploration-like scenario with various materials. Experiment results demonstrate that the proposed system can create useful material maps in unknown environments.


\end{abstract}

\section{Introduction}
\label{sec:introduction}

The utility of robots in exploration and mapping applications has gained significant interest and advancements in recent years. Mobile robots are used in a variety of situations such as search and rescue scenarios, firefighting aids, service and logistics, domestic aids, and more. Nevertheless, numerous unsolved problems remain when deploying robots in unknown environments. In particular, it is important for the robot to perceive and learn the properties of an unknown environment to increase its autonomy and effectively execute its mission \cite{katz2014perceiving}. For instance, in search and rescue operations, it is essential to know the location of doors and other access points to preplan the operation.

To address this problem, researchers have utilized several sensing modalities and machine learning algorithms to classify different objects and materials. Of these modalities, computer vision has become prominent because of the availability of public image datasets and recent advances in deep learning algorithms \cite{wieschollek2016transfer}. There are several search and rescue robots in use that depend on image processing and vision-based techniques \cite{kim2016feature,matsuno_rescue_2004}; however, they tend to fail when the lighting conditions are bad  or the environment is smoky which significantly reduces the visibility of the scene. 

Tactile and acoustic sensing techniques that are robust to poor lighting conditions have been well studied and proven effective in similar applications \cite{hoelscher2015evaluation,lopez2017classification}. While tactile sensors alone require a variety of contact motions and potentially lengthy contact duration with the surface material, a combination of tactile and acoustic (sound) signals reduces the complexity when employing a simple interaction such as tapping. In fact, elucidating the properties of a target material through machine analysis of sounds generated from it is a well-studied topic. In particular, studies using the sound of tapping to identify material type date back to the work of Durst and Krotkov in 1995 \cite{durst1995object} where peaks in the frequency domain were used for the classification. 

Recently, the authors in \cite{lopez2017classification} used NAO humanoid robots to manipulate target objects (picking up and forcefully hitting it) and used the dominant frequency of the recorded sounds to classify the objects. Similar work was performed in \cite{torres2005tapping}.
In \cite{sinapov2008interactive}, using the Fourier analysis of the sounds resulting from a robot manipulator performing several actions on the target object (grasp, push, or drop), the authors can accurately ($\approx$ 97\%) classify up to 18 objects with a Bayesian classifier. Motion aided audio signal analysis has also been used to detect touch gestures \cite{alonso2017detecting}, and terrain and surface types  \cite{roy1996surface}. Moreover, the integration of such sound-based analysis to robot exploration is still an evolving research area. For instance, in \cite{watanabe2015robot}, the authors used tapping sounds along with a LiDAR scan from a mobile robot to create a map of the impact locations for assisting the human inspector during hammer sounding inspections of the concrete walls and buildings. 

\begin{figure}[t] 
  \centering
  \includegraphics[width=0.95\columnwidth]{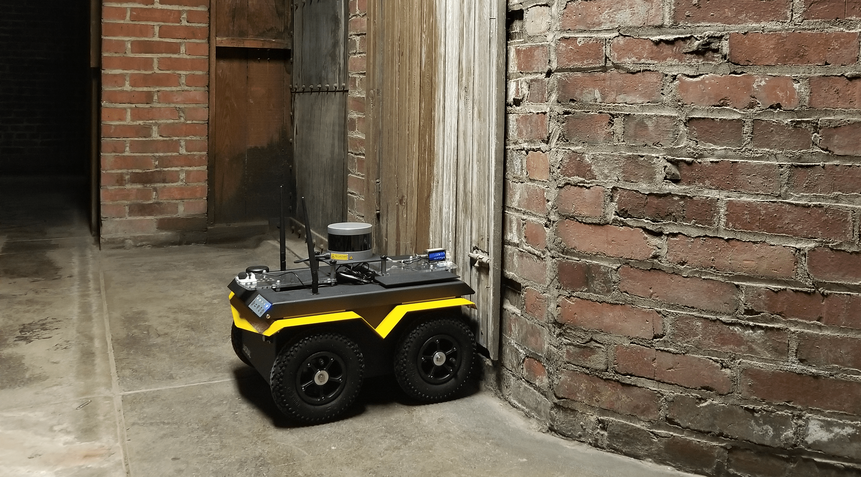}
   \caption{A robot autonomously exploring and tapping the objects in order to perceive its environment.}
   \label{fig:robot_tapping}
\end{figure}

Although a lot of tactile and acoustic sensing methods research have been proposed in literature, a very few have addressed the complete autonomous mapping systems that can label objects on a map according to their material types using their methods. Therefore, in this paper we propose a novel autonomous material mapping system that enables a robot to explore an unknown environment and at the same time identify surrounding materials in that environment along with their locations. More specifically, the goal of this work is to develop an exploration strategy using SLAM and integrate it with a solenoid based tapping mechanism to detect the constituent materials of the objects present in the environment through machine learning algorithms. This work being motivated from search and rescue scenarios which are usually time constrained, the challenge is to fasten the exploration process. Fig.~\ref{fig:robot_tapping} depicts our robot exploring an unknown environment and tapping on the objects present there to identify their constituent materials.

The main contributions of our work are two-fold:
\begin{itemize}
    \item We introduce and validate a simple and effective neural network for classifying the materials based on the sounds produced from tapping various objects.
    \item We propose an efficient exploration strategy for a mobile robot to autonomously explore an unknown environment and identify the constituent materials of the objects in the environment in a short time and hence, creating a material map. 
\end{itemize}

\section{Overview of the Proposed System}
\label{sec:proposed_sys}
The proposed system identifies the materials in an unknown environment and creates a material map which localizes the various materials in the environment over the occupancy grid map of that environment. The system consists of two key components: 1) the material classification system, and 2) the exploration system both implemented on mobile robot. The mobile robot consists of a linear solenoid, a dual microphone set, and a LiDAR. The solenoid is used for tapping the object, the dual microphone set is used to record the sound produced by the tap of the solenoid, and the LiDAR is used for the localization, mapping and navigation.  The complete information about the mobile robot platform and its design is elaborated in Section \ref{sec:robo_platform}.

The material classification system enables the identification of the materials through the tapping sound generated using the solenoid on the robot. The robot taps on various objects and the sounds generated is recorded and processed to identify the material  corresponding to the object. The system uses Mel-Frequency Cepstral Coefficients (MFCC) as features coupled with a convolutional neural network (CNN) to classify the material based on the tapping sounds. The complete details on the implementation of the material classification system is  explained in Section \ref{sec:material_class}.

\begin{figure}
  \centering
  \includegraphics[width=0.95\columnwidth]{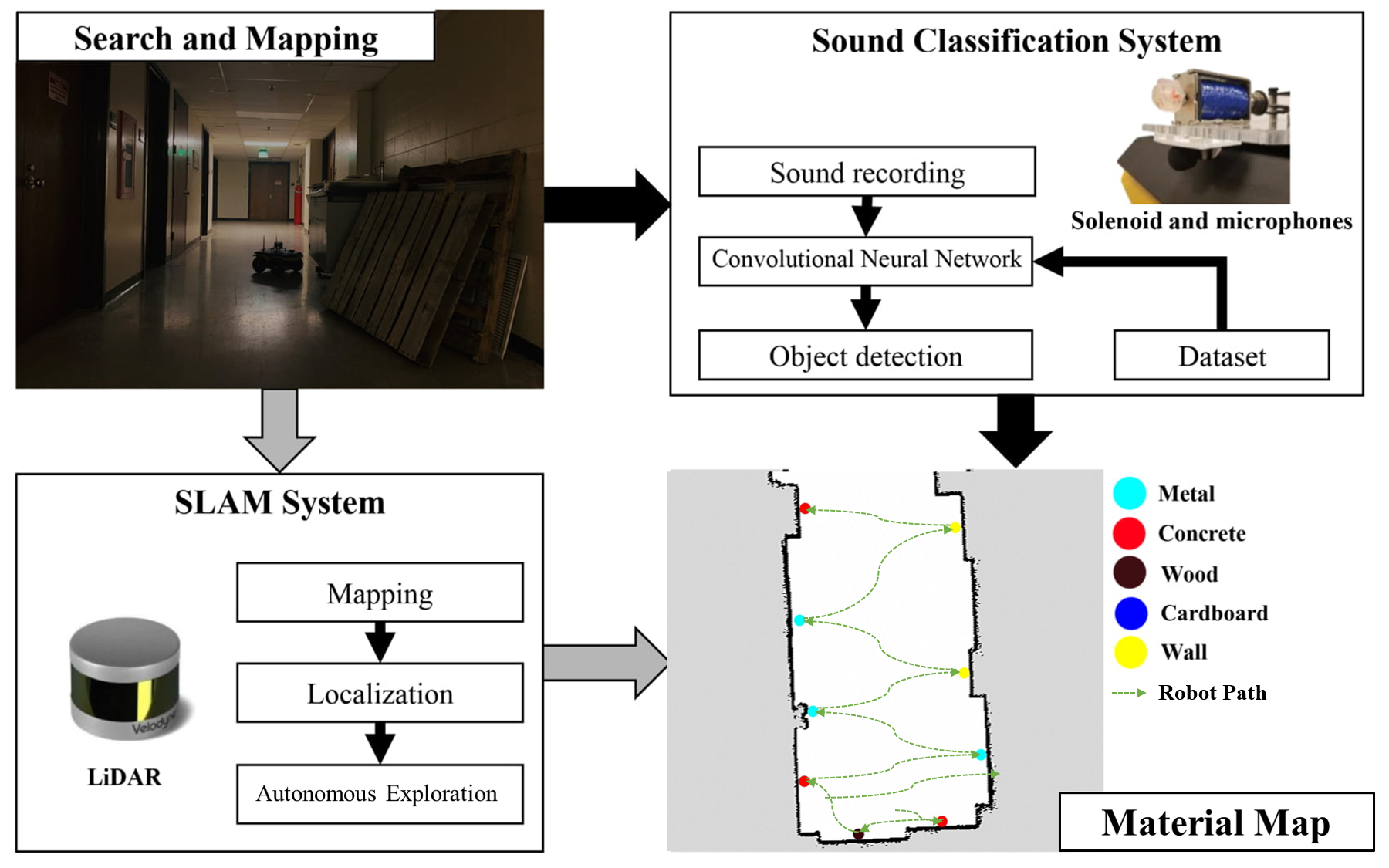} 
    \caption{Structure and flow of the proposed autonomous exploration and material mapping system working together to build the material map.}
   \label{fig:architure}
\end{figure}

The autonomous exploration system assists the robot in autonomously exploring and identifying the materials in an environment. It relies on the information provided by the SLAM module (GMapping \cite{gmapping}) to map and to navigate. The initial map (partial map of the environment) built by the SLAM module is processed geometrically to identify the points at which the robot needs to tap and identify the materials. The points to be sampled are selected such that robot can identify all the materials in the region with the least number of taps. The number of taps is minimized as an effort to reduce the total exploration time. The robot then moves to these points one after the other and taps using the solenoid. The tapping sound produced is recorded and then processed by the material classification system to detect the material. The materials identified are overlaid on the occupancy grid map as markers. The robot then uses frontier-based exploration \cite{Yamauchi1997AFA} to explore the environment further. The robot then process the map of the new region explored and identifies the materials in it by tapping. This process is iteratively repeated until the entire environment has been explored and the materials in the environment also have been identified. The completed details about the functioning of the exploration system are described in Section \ref{sec:exploration}. The overall structure and the flow of the proposed system are illustrated in the Fig. \ref{fig:architure}.

\section{Robot Platform}
\label{sec:robo_platform}
This section describes the design and hardware configuration of the mobile robot platform used for the exploration and the mapping of the materials. The section also elaborates on the active noise reduction system which is an integral part of the robot platform.

\subsection{Hardware Configuration}
\begin{figure}
  \centering
  \includegraphics[width=0.9\columnwidth]{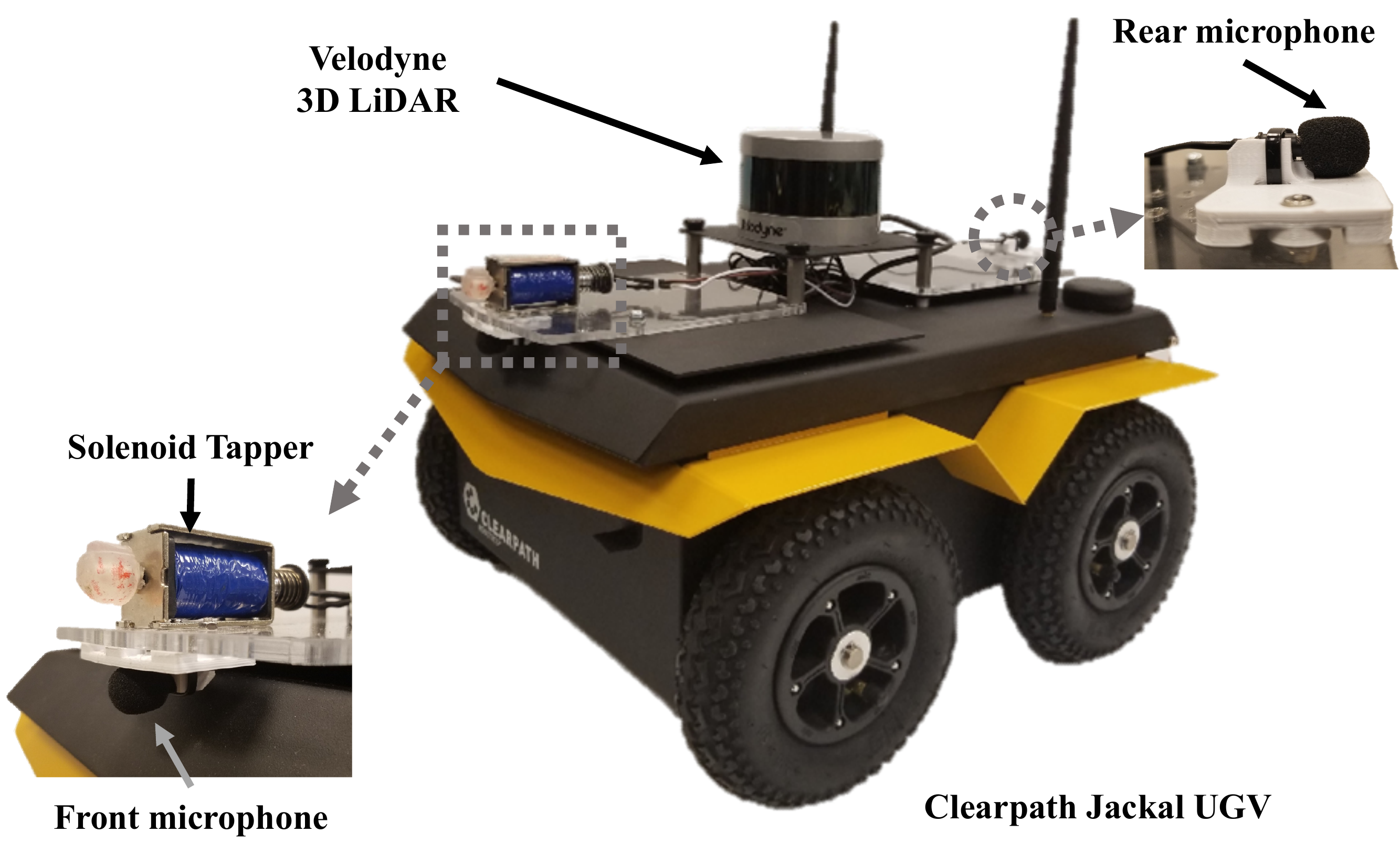}
   \caption{Hardware setup of the robot platform used.}
   \label{fig:hardware_elements}
\end{figure}

\begin{figure}
\centering
    \begin{subfigure}{0.49\linewidth}
        \includegraphics[width=\linewidth]{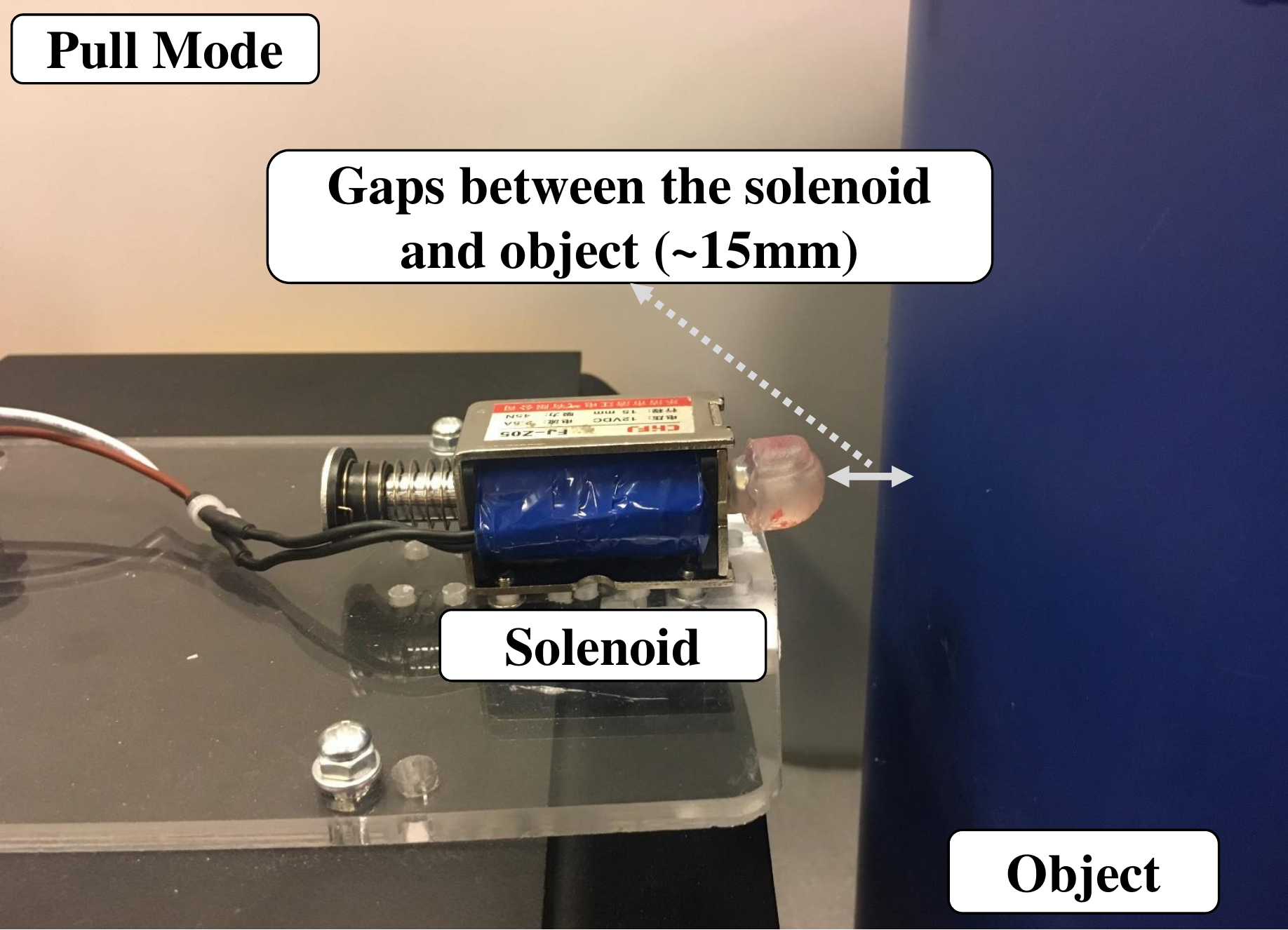}
        \caption{Pull Mode}
        \label{fig:operation_solenoid_a}
    \end{subfigure}\hfill
    \begin{subfigure}{0.49\linewidth}
        \includegraphics[width=\linewidth]{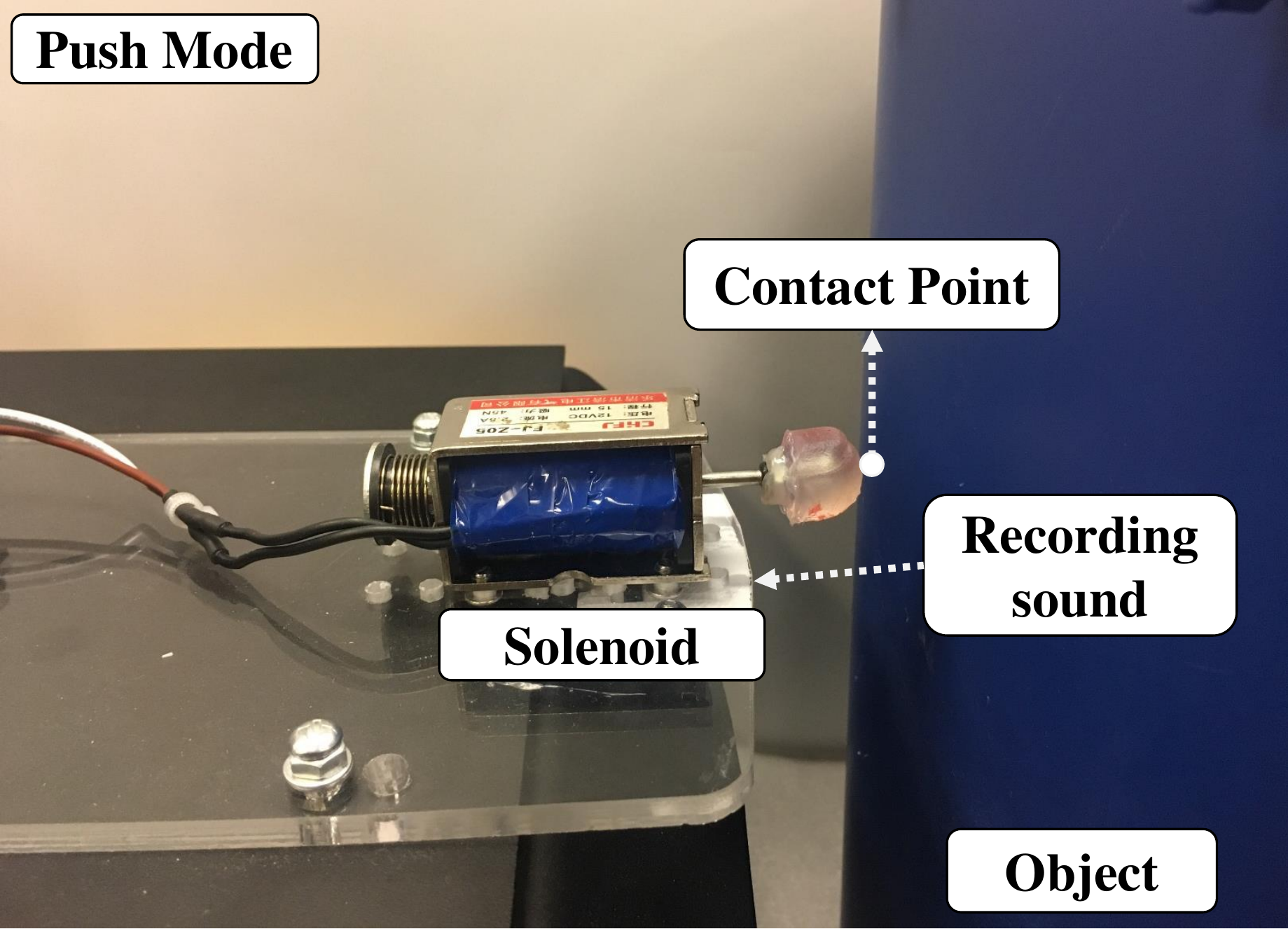}
        \caption{Push Mode}
        \label{fig:operation_solenoid_b}
    \end{subfigure}
\caption{Operation of the solenoid to generate a tap sound. }
\label{fig:operation_solenoid}    
\end{figure}

The robot was built on a Jackal UGV from Clearpath Robotics (it is worth noting that although Jackal UGV was used in this paper, the design is generalized such that it could be implements on many ground robot). The material mapping robot consists of three key components: a LiDAR, a linear solenoid, a dual microphone set. The primary use of the LiDAR is to map the environment and also guide in the autonomous exploration. A Velodyne VLP-16 was used in our robot (though a 3D LiDAR was used, but only 2D data was used). A linear solenoid switch that can be controlled to extend/retract was used to tap the various materials in the environment. The overall hardware setup of the material mapping is shown in Fig. \ref{fig:hardware_elements}, and the working of the solenoid in pull (retract) and push (extend) modes is depicted in Fig. \ref{fig:operation_solenoid}. The tip of the linear solenoid was covered with a plastic cap to make the solenoid tip acoustically compatible as the elasticity of the tip plays an important factor \cite{SCHENKMAN1986}. The solenoid used had a stroke length of about 15 $mm$ and applies a force of about 45 $N$. The force of the solenoid is good enough to produce sound but not too large that it can cause damage to the environment. 

In addition to the LiDAR and the linear solenoid, the material mapping robot is equipped with a dual microphone set. The microphones are used to record the sound produced by the tap of the solenoid. Among the two microphones, one (on the front side of the robot) is placed closer to the solenoid tapper, so that it obtains the tapping sound with high sound clarity. The other microphone (on the rear side of the robot) is placed away from the solenoid tapper to capture the noise in the environment (including the noise sounds created by the robot's movement) with less influence of the tapping sounds. This system is later used to reduce the impact of the environmental noise on the material classification.

\subsection{Active Noise Reduction}
Another critical component of the the robot platform is active noise reduction. To deploy the material mapping robot in the real-world, the recording system needs to remove or reduce the ambient acoustic noise for accurate analysis. The synchronized sound signals from the front microphone (tapping sound) and the rear microphone (background noise) are sent to the adaptive filter, which efficiently matches the unknown noise characteristics with a Finite Impulse Response (FIR) model and applies the error correction through Normalized Least Mean Squares (NLMS) algorithm that is generally used in signal enhancement algorithms. The anti-noise is generated by inverting the FIR output and then combined with the signal from the front microphone to effectively remove the background noise \cite{NLMS_theory}. This processed audio signal is later used in the material classification system.

\section{Material Classification using Tapping Sound}
\label{sec:material_class}
In this section, we describe a tapping sound-based material classification system using Convolutional Neural Networks (CNN) along with the dataset creation and the classification results.

\subsubsection{Selection of Classes and Dataset Creation} 
The development of a classification model requires a reference dataset based on which the model could be developed. To the best of our knowledge, there are no publicly available datasets consisting of tapping sounds of various materials. Hence, we created a dataset with an extensive collection of tapping sounds from various objects, such as wood, metal, glass, plastic, cardboard, wood, concrete, and wall (hardboard). The materials were selected based on a visual survey for the common materials found in everyday life. The tapping sounds from objects like trash bins, storage cabinets, wall, cardboard boxes, doors, tables, and so on which are made with these materials (as their principal composition) were recorded to build the dataset. The tapping sounds were recorded using the robot setup described in the previous section. To ensure that the various frequencies corresponding to the same material are captured, objects of various sizes were used and they were tapped at various locations. We also included a class consisting of the empty tap sounds (with no target object) to account for tapping issues such as the solenoid tip unable to contact the target material. So the total number of classes in our experiments is eight (seven materials + one empty). The dataset consists of 1,045 tapping sounds in total with an average of around 100 sample sounds per class. 

\subsubsection{Feature Extraction} 
\begin{figure*}
\centering
\includegraphics[width=0.97\textwidth]{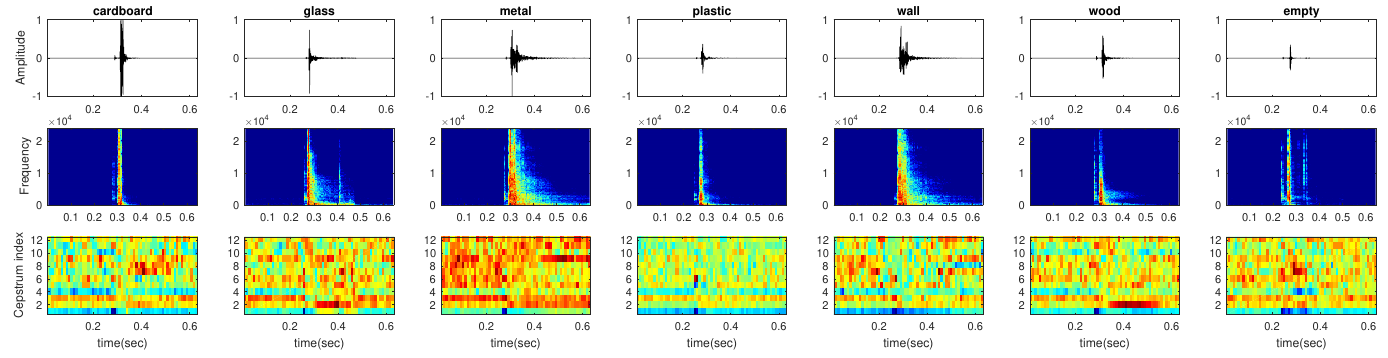}
\caption{Tapping sound signal characteristics in time domain (amplitude), Fourier domain (frequency), and cepstral domain (MFCC indices) are shown for the following materials (in the order from left to right): cardboard, glass, metal, plastic, wall, wood, and empty (background noise). }
\label{img:all_material_sound_analyzing}
\end{figure*}

Mel-Frequency Cepstral Coefficients (MFCC) \cite{mogran2004automatic} were used as the feature for the material classification. MFCC were used since they summarize the frequency distribution making it feasible to evaluate both the frequency and time characteristics of the sounds. 40 MFCC values for 45 frames were extracted and used. Also, from Fig. \ref{img:all_material_sound_analyzing}, it can be seen that Mel-frequency Cepstral Coefficients are very distinct for every material compared to spectrogram-based features as used in \cite{strese2017multimodal,fujisaki2015perception}. This distinct nature of the MFCC enables the construction of a robust classification model.  

\subsubsection{Convolutional Neural Network classifier}
\begin{figure}
\vspace{-0.2cm}
\centering
\includegraphics[width=0.42\textwidth]{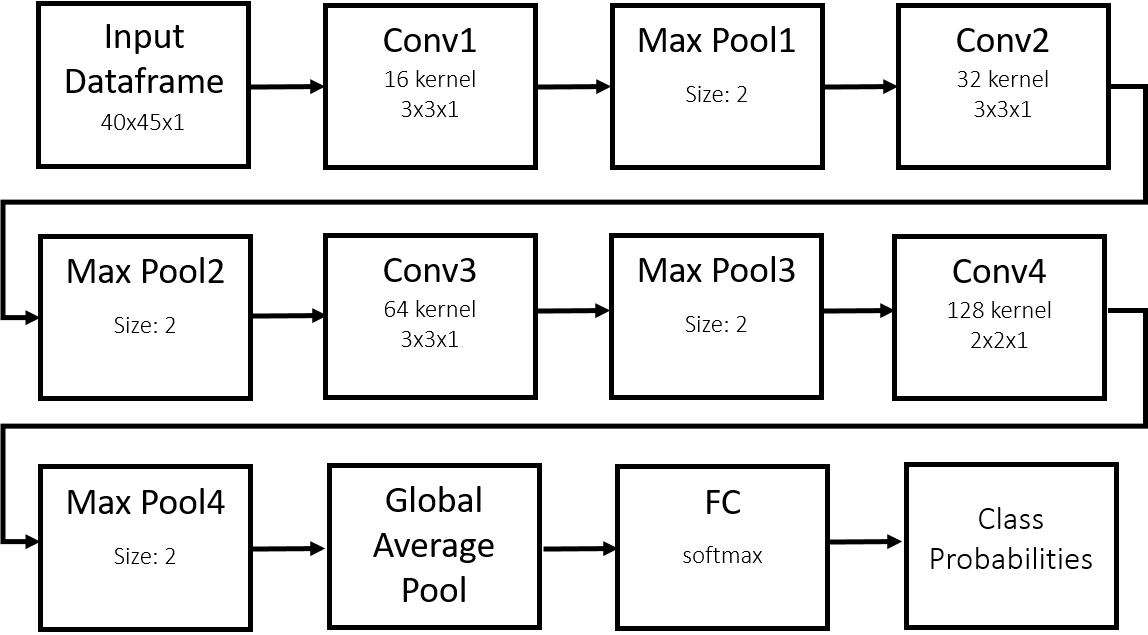}
\caption{Block diagram illustrating the neural network architecture used for classifying the tapping sounds. Each block corresponds to a layers in the CNN. The kernel sizes of the layer and the other parameters such as filter size and activation are presented within the blocks.}
\label{img:cnn_archi}
\end{figure}

Convolutional Neural Networks (CNN) were used for the tapping sound classification. An input vector of size 40 $\times$ 45 was fed to the network. This vector corresponds to 40 MFCC values across 45 frames from the audio signal. The classifiers estimate the probabilities for various materials. The CNN model uses four convolution layers of kernel sizes 16, 32, 54, and 128. The first three layers had a filter size of $3\times3$ and the last layer had a filer size of $2\times2$. Rectified Linear Unit (ReLU) activation was used for all the four layers. The convolution layers were followed by a global average pooling layer aimed at reducing the dimensions which was followed by a fully-connected layer with softmax as activation. The flow and the various blocks of the neural network architecture along with the filter sizes and other parameters have been depicted in Fig. \ref{img:cnn_archi}.

\subsubsection{Classification results}
\begin{figure}
\vspace{-0.7cm}
  \centering
    \includegraphics[width=0.91\linewidth]{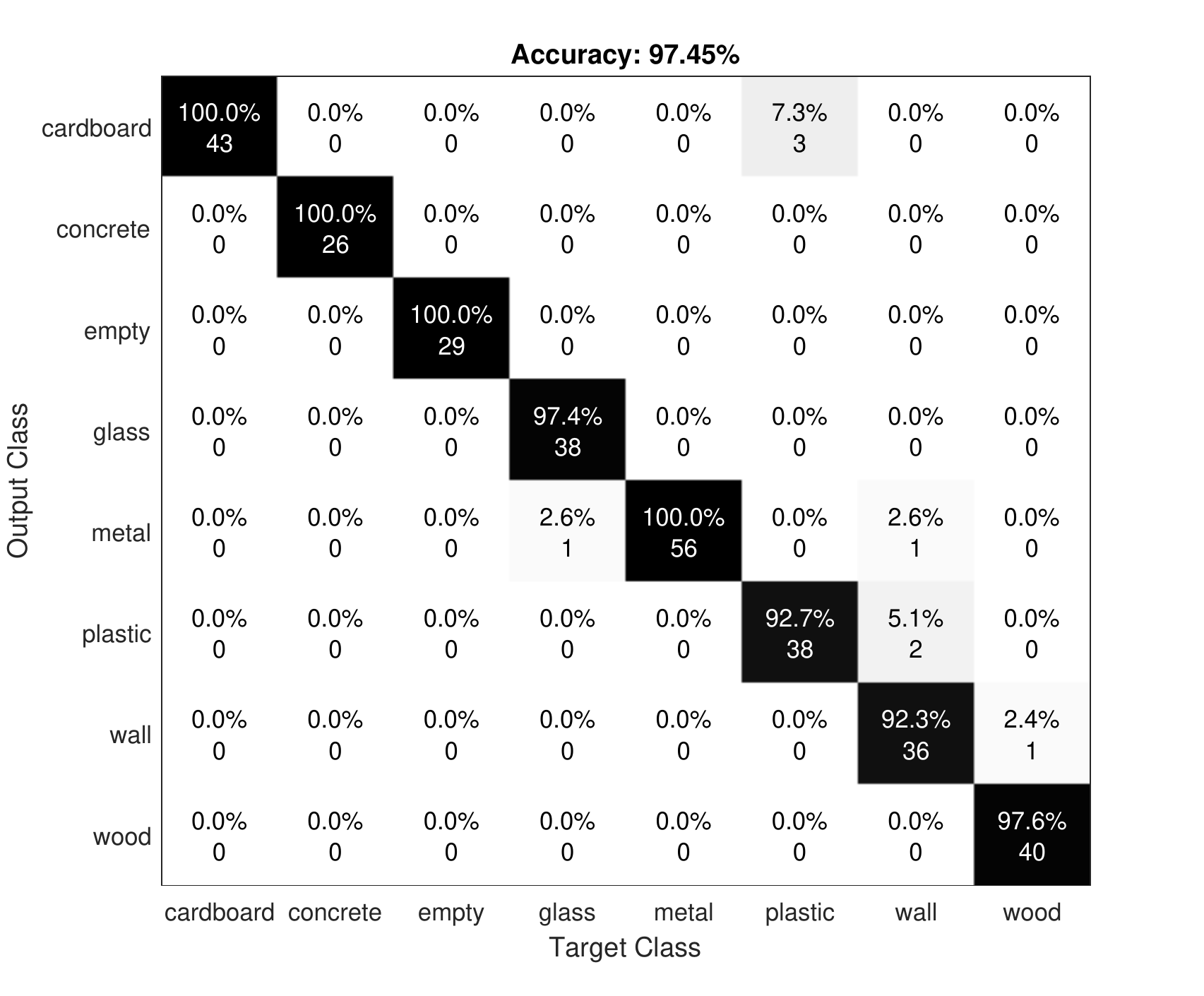}
  \caption{The confusion matrix of our CNN classifier on the test data. The classifier has an accuracy of 97.45\%.}
  \label{img:result_confusion_matrix}  
\end{figure}

The dataset created was split into train and test sets in the ratio of 70:30. A CNN classifier was trained and then was validated on the test set. The normalized results of the classification on the test set are presented as a confusion matrix in Fig. \ref{img:result_confusion_matrix}. The mean accuracy of the classification results is 97.45\%. From the confusion matrix, it can be observed that there is a reasonable misclassification between plastic and wall, and plastic and cardboard. These are due to the similarities in the tapping sound for these sets of materials. In addition to the CNN classifier, we also applied an SVM based classifier, which resulted in a mean accuracy 91.58\% but the CNN classifier significantly outperformed it especially in cases of poorly classified materials such as plastic using the SVM.

\section{Autonomous Exploration}
\label{sec:exploration}
This section describes the iterative exploration strategy used by the robot to build the material map. The exploration system aids in the extraction of the points to tap and in the navigation. The system uses gmapping SLAM \cite{gmapping} for the mapping and localization of the robot along with the ROS navigation stack for the navigation. The exploration system functions by iteratively sampling the objects in the robot's vicinity and exploring unknown regions until the entire environment has been mapped. 

\subsection{Identification of Tapping Points}
Once the robot is deployed into the site, it starts mapping that environment using SLAM. This creates an initial map covering the area around the robot (area covered depends on the range of the LiDAR). Now, this map information is used to identify the points to tap in the region mapped so far. The points to be sampled by the robot should lie on the border between the empty (or free) and occupied regions (white and black pixels in the occupancy grid). Hence, the boundary points need to extracted to facilitate the computation of the tapping points ($tap\_points$ in the Algorithm \ref{alg:pseudo}). The boundary points ($boun\_pts$ in the Algorithm \ref{alg:pseudo}) are obtained looking at the eight neighboring pixels of all the occupied pixels. If a pixel is surrounded by at least one pixel corresponding to the free space (white pixel), then that pixel is considered to be a part of the boundary.

A boundary graph ($boun\_graph$ in the Algorithm \ref{alg:pseudo}) is constructed using the various boundary pixels and with neighboring pixels forming edges. The connected components ($con$ in the Algorithm \ref{alg:pseudo}) in the graph are computed and they give the various disjoint boundary segments on the map. There can be many small segments present, which usually correspond to noise or partially mapped boundaries in unexplored regions. The segments whose length is less than the minimum segment length, $\sigma$ are removed and are not considered for the tapping point estimation. In the implementation,  $\sigma$ of 25 points was used.

The segments obtained may correspond to a single object or multiple objects. In most cases, the presence of multiple objects in one single segment creates concavities in the boundaries. For instance, when a box is placed close to the wall, it creates a concavity at the points where the box is in contact with the wall. Now, the map is further segmented in sub-segments based on its geometry by finding the corner points (convex and concave). Shi-Tomasi corner detection algorithm \cite{shi1994good} was used to find these corner points ($cor$ in the Alg.~\ref{alg:pseudo}). Creating segments based on corners yields boundaries corresponding to each side of an object.

Now, if the robot makes one tap per segment it should ideally capture all the materials in that region. But sometimes, the boundaries between two different objects are not marked by a change in curvature. For instance, the doors and walls might be at the same level and they form one big straight line, which will be recognized as one single segment. Performing one tap on this will not enable us to identify all the materials. Hence, sometimes we need to perform more than one tap for certain segments.

In order to achieve this without having the need to tap each and every point, a human-defined distance parameter, $\gamma$ is introduced. The distance parameter, $\gamma$ is the interval at which the taps need to be done. In other words, $\gamma$ is the distance between two successive taps within a segment. If the length of a segment is less than 1.5 times $\gamma$, the midpoint of that segment is identified as the tapping point (e.g. the sides of a box, a trash can and so on). In the case of bigger segments, multiple tapping points are selected which are separated by a distance $\gamma$ (e.g. longer segments of the wall). This parameter enables us to keep the balance between the time taken for the exploration and the number of materials detected. 

Once all the tapping points are estimated, the orientation the robot needs to maintain at that point for a successful tap is computed by finding the perpendicular to the direction of the boundary at that point. The robot orienting itself perpendicular to the object surface enables the solenoid to be in complete contact with the surface during the tap. 

Now, that the tapping points with the tap directions are estimated, the robot plans the shortest path through all these points using traveling salesman problem (TSP) and the robot moves from point to point, tapping and identifying the materials at all those points. As the robot identifies the material they are overlaid on the occupancy grid map and hence constructing the material maps. 

\subsection{Frontier-based Exploration}
Once the robot finishes tapping and sampling the points around it, the robot needs to explore the other unexplored regions and identify the materials there. To realize this continuous exploration, we use the frontier exploration algorithm \cite{Yamauchi1997AFA}. The closest boundary between the explored regions and the unexplored region is identified and the robot navigates to that boundary. Moving to this boundary facilitates the robot to map new unknown regions of that environment.

Now, the extended map is found by removing the parts of the map already explored. The extended map is processed using the tapping point identification methods elaborated before. During the process of exploration, some of the segments computed in the previous iteration might be extended. In those cases, the new tapping points on those segments are estimated based on the points already tapped on that segment. 

This process of tapping and exploration is repeated until there is not any boundary between explored and unexplored regions (i.e. the entire environment has been mapped and sampled for materials too). The entire process comes to a halt when the robot has successfully explored the entire region. In this process of iterative exploration, the path from one tapping to another is optimized locally on the extended map using TSP and not globally. The pseudo-code for the entire exploration system is presented in Algorithm \ref{alg:pseudo}.

\begin{algorithm}[t]
\caption{Pseudo-code for the autonomous exploration}
\begin{algorithmic}[1]
\Require $\sigma$, minimum segment length
\Require $\gamma$, the distance parameter
\State $map\gets current\; occupancy\; grid$
\State $previous\_map\gets null$
\State $boundary\_segments\gets [ ]$
\State $tap\_points \gets [ ]$
\While {$frontier \neq null$}
    \State $occupied\_pts \gets occupied\; pixels\; in\; the\; map$
    \State $free\_pts \gets free\; pixels\; in\; the\; map$
    \State $boun\_pts \gets occupied\_pts\; neighboring\; free\_pts$
    \State $boun\_graph \gets graph\; constructed\; using\; boun\_pts$
    \State $con \gets connected\;components\;in\;boun\_graph$
    \ForAll {$c \in \mathcal \;con $}
         \If {$length\_of(c) > \sigma$}
            \State $cor \gets corner\; points\;in\;c$
            \State $seg \gets segments\;from\;breaking\;c\;at\;cor $
            \State $boundary\_segments.append(seg)$
         \EndIf
    \EndFor
    \ForAll {$s \in \mathcal \;boundary\_segments $}
        \If {$length\_of(s) > 1.5*\gamma$}
            \State $tap\_points.append(points\;in\;s\;at\;interval\;\gamma)$
        \Else
            \State $tap\_points.append(midpoint\;of\;s)$
        \EndIf
    \EndFor
    \State compute the direction for each tap point
    \State navigate and sample all the points in$\;tapping\_points$
    \State $frontier \gets closest\;frontier\;boundary\;point$
     \If{$frontier \neq null$}
        \State navigate to the frontier
        \State $previous\_map \gets map$
        \State $map\gets current\; occupancy\; grid$
        \State $map\gets map\;-\;previous\_map$
    \EndIf
\EndWhile
\State \textbf{end}
\end{algorithmic}
\label{alg:pseudo}
\end{algorithm}

\begin{figure*}[t]
\centering
    \begin{subfigure}{0.195\linewidth} 
        \includegraphics[width=\linewidth]{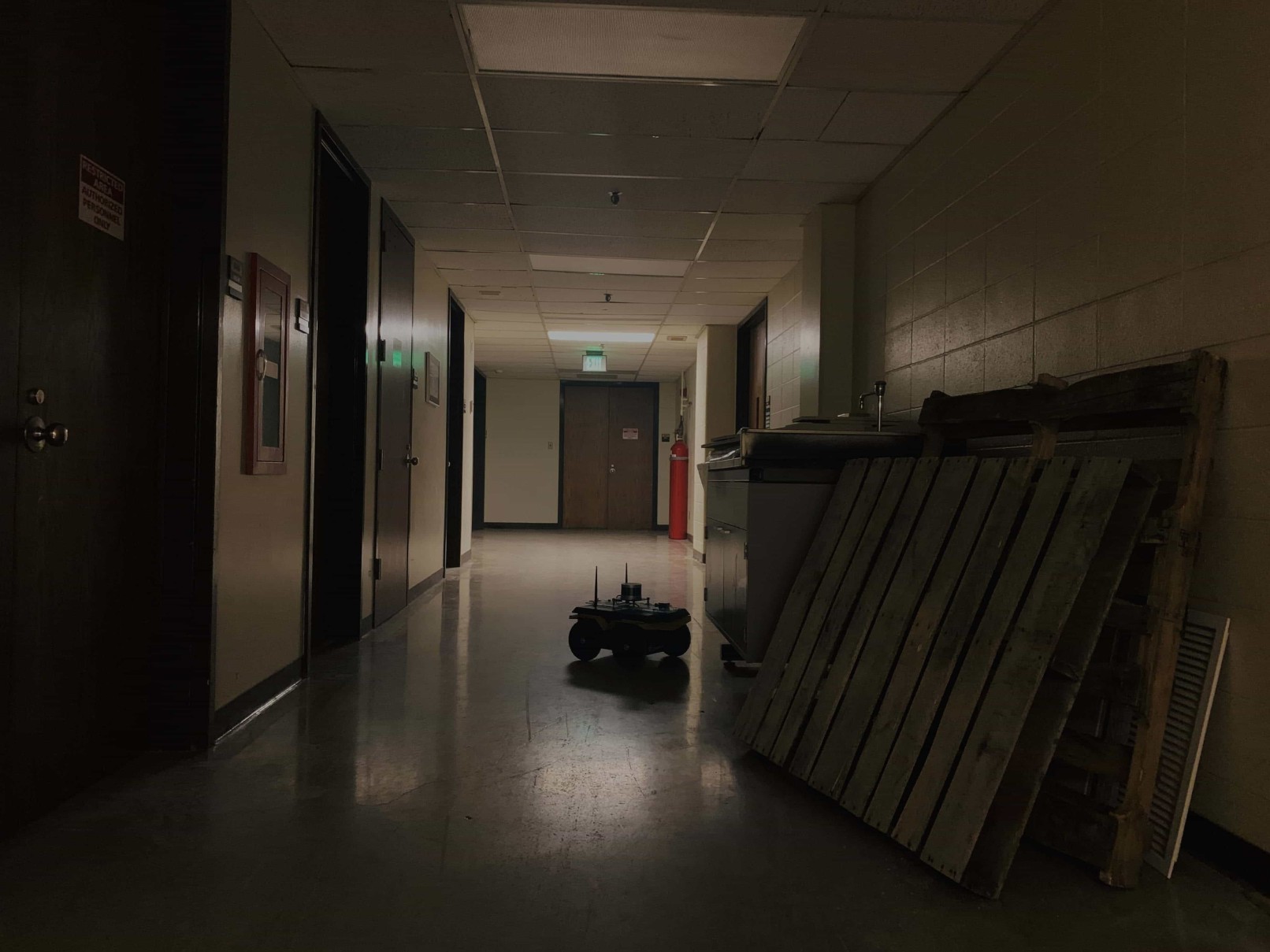}
        \caption{}
        \label{fig:location_a}
    \end{subfigure}\hfill
    \begin{subfigure}{0.195\linewidth} 
        \includegraphics[width=\linewidth]{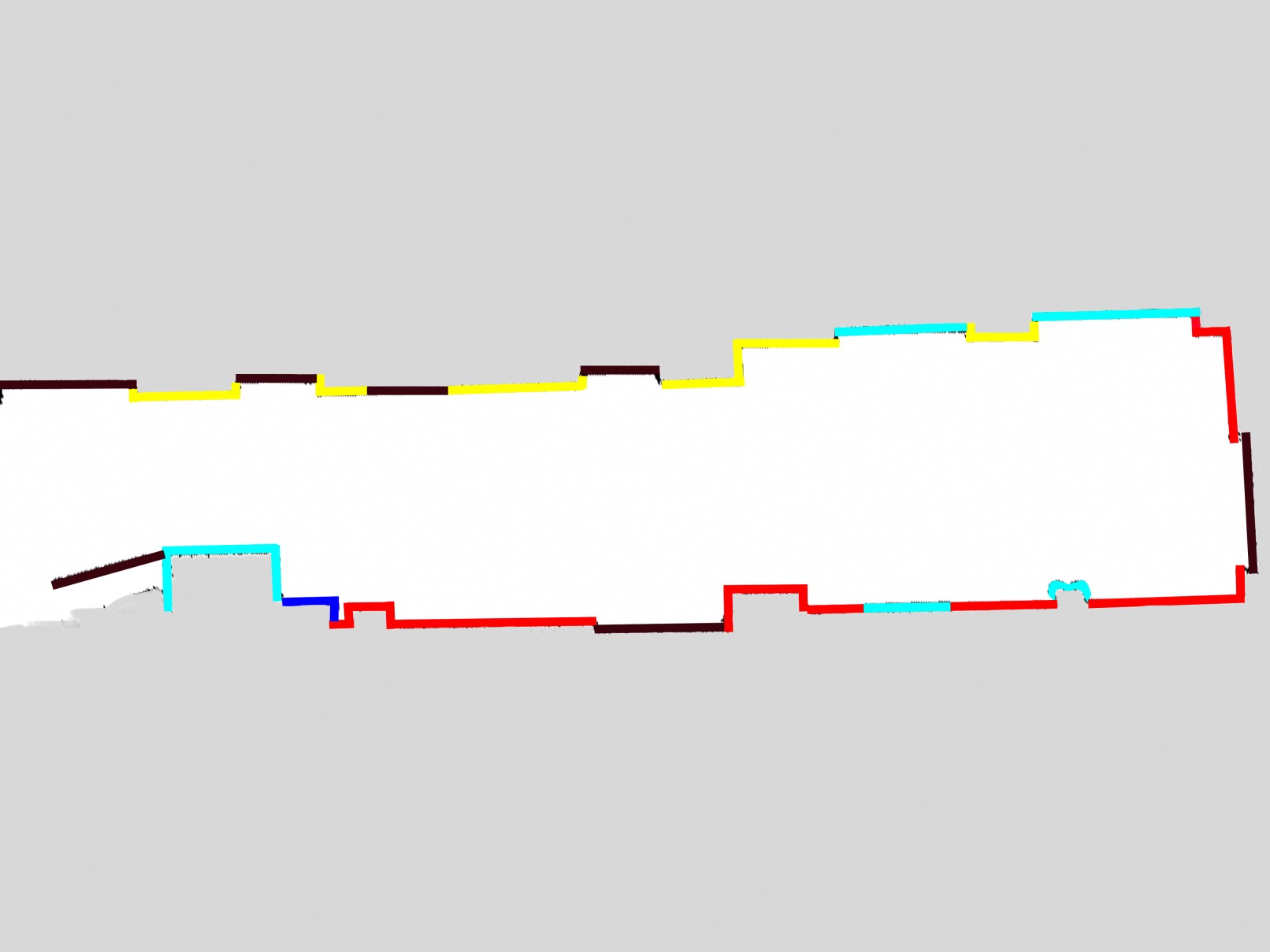}
        \caption{}
        \label{fig:loc_a_gt}
    \end{subfigure}\hfill
    \begin{subfigure}{0.195\linewidth} 
        \includegraphics[width=\linewidth]{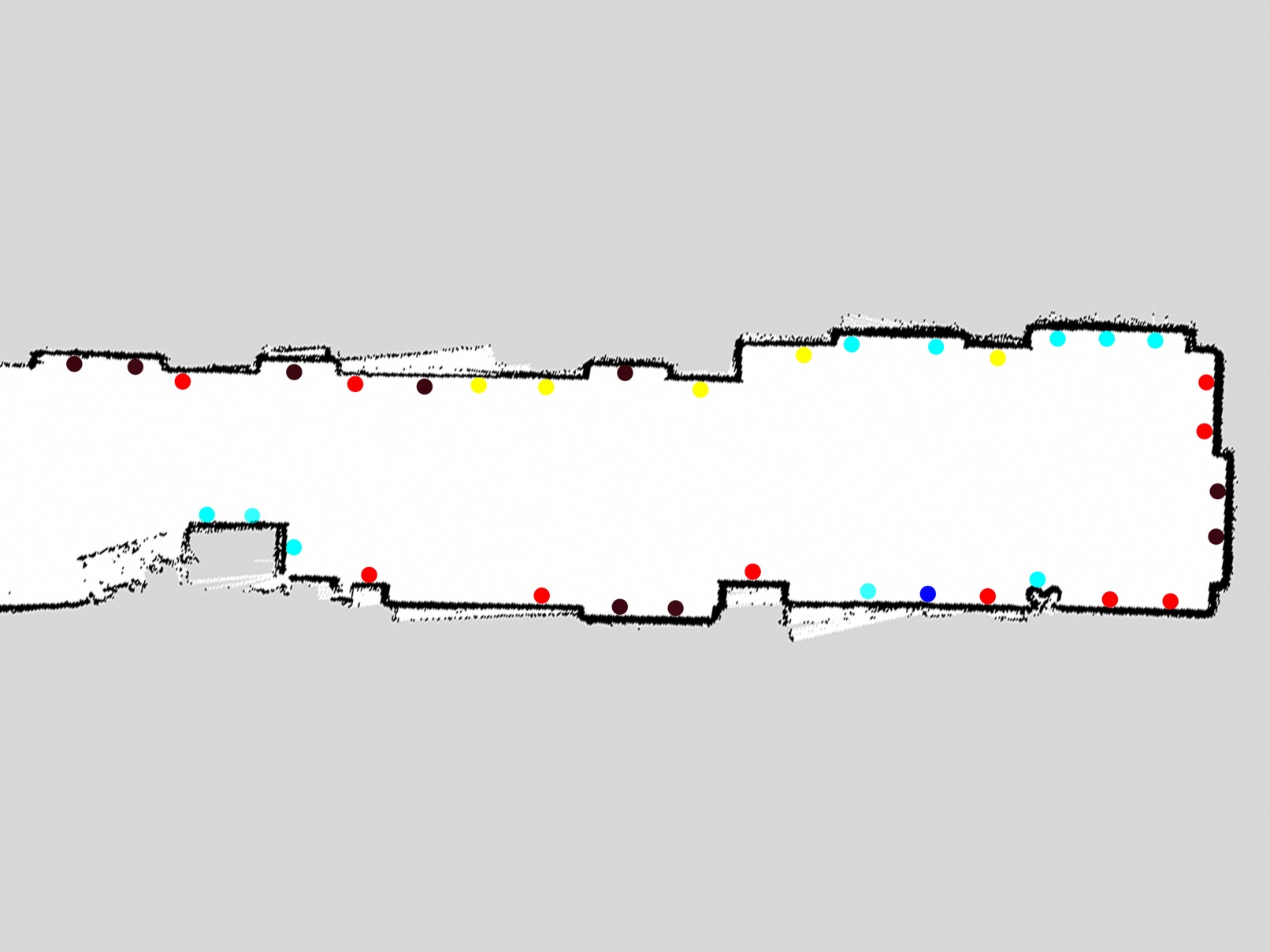}
        \caption{}
        \label{fig:loc_a_1}
    \end{subfigure}\hfill
    \begin{subfigure}{0.195\linewidth} 
        \includegraphics[width=\linewidth]{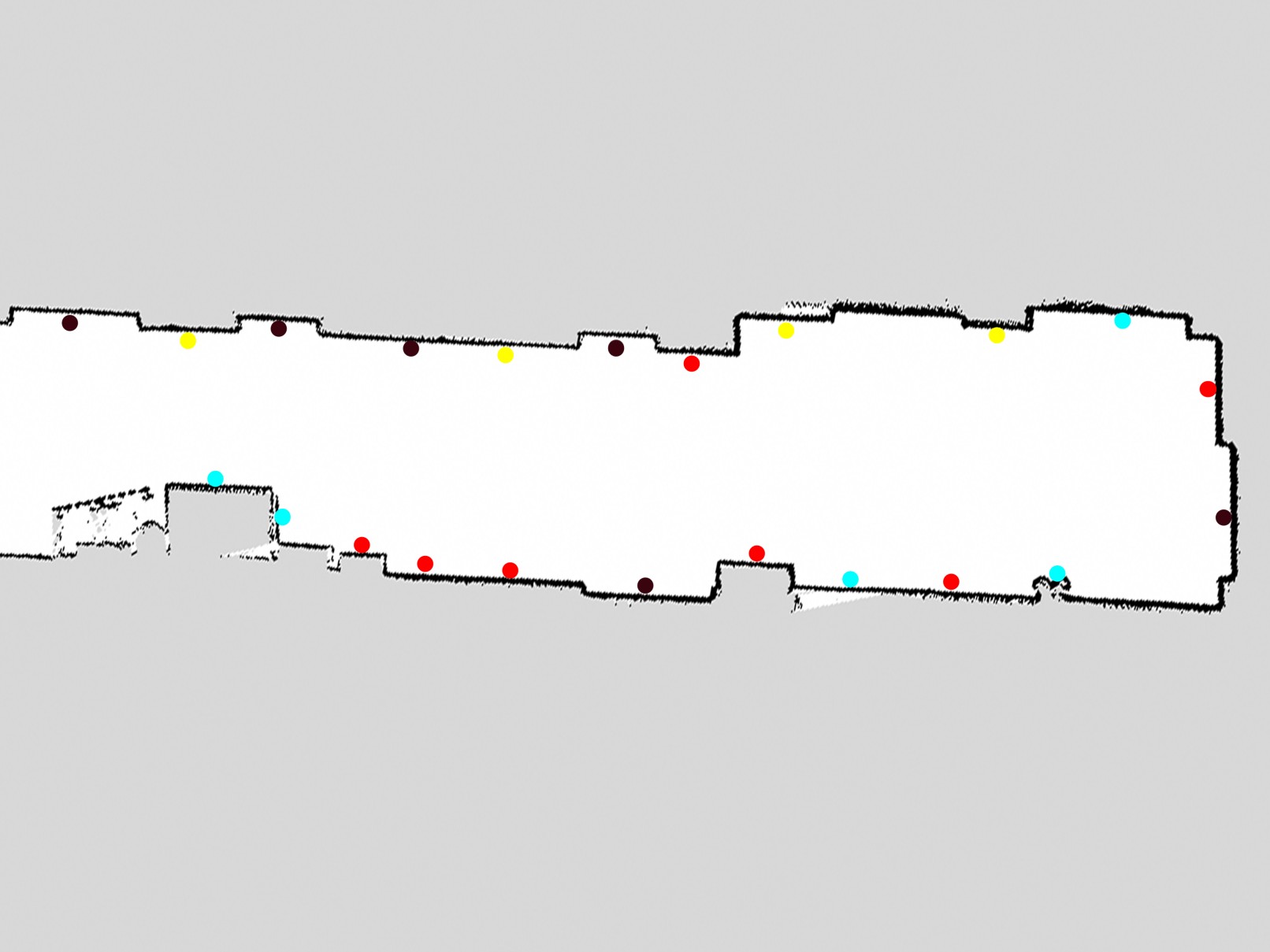}
        \caption{}
        \label{fig:loc_a_2}
    \end{subfigure}\hfill
    \begin{subfigure}{0.195\linewidth} 
        \includegraphics[width=\linewidth]{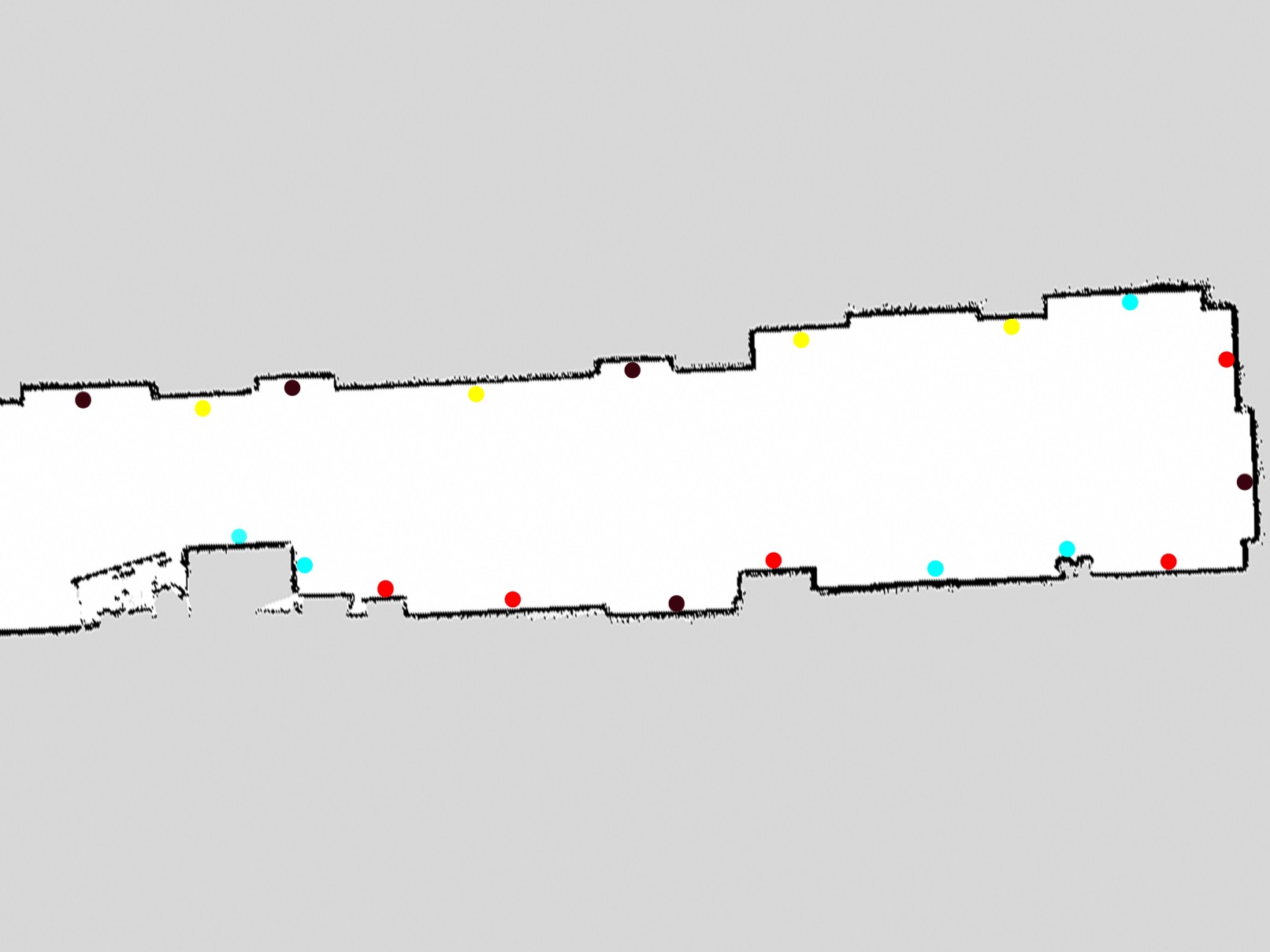}
        \caption{}
        \label{fig:loc_a_3}
    \end{subfigure}\hfill
    \vfill
    \begin{subfigure}{0.195\linewidth} 
        \includegraphics[width=\linewidth]{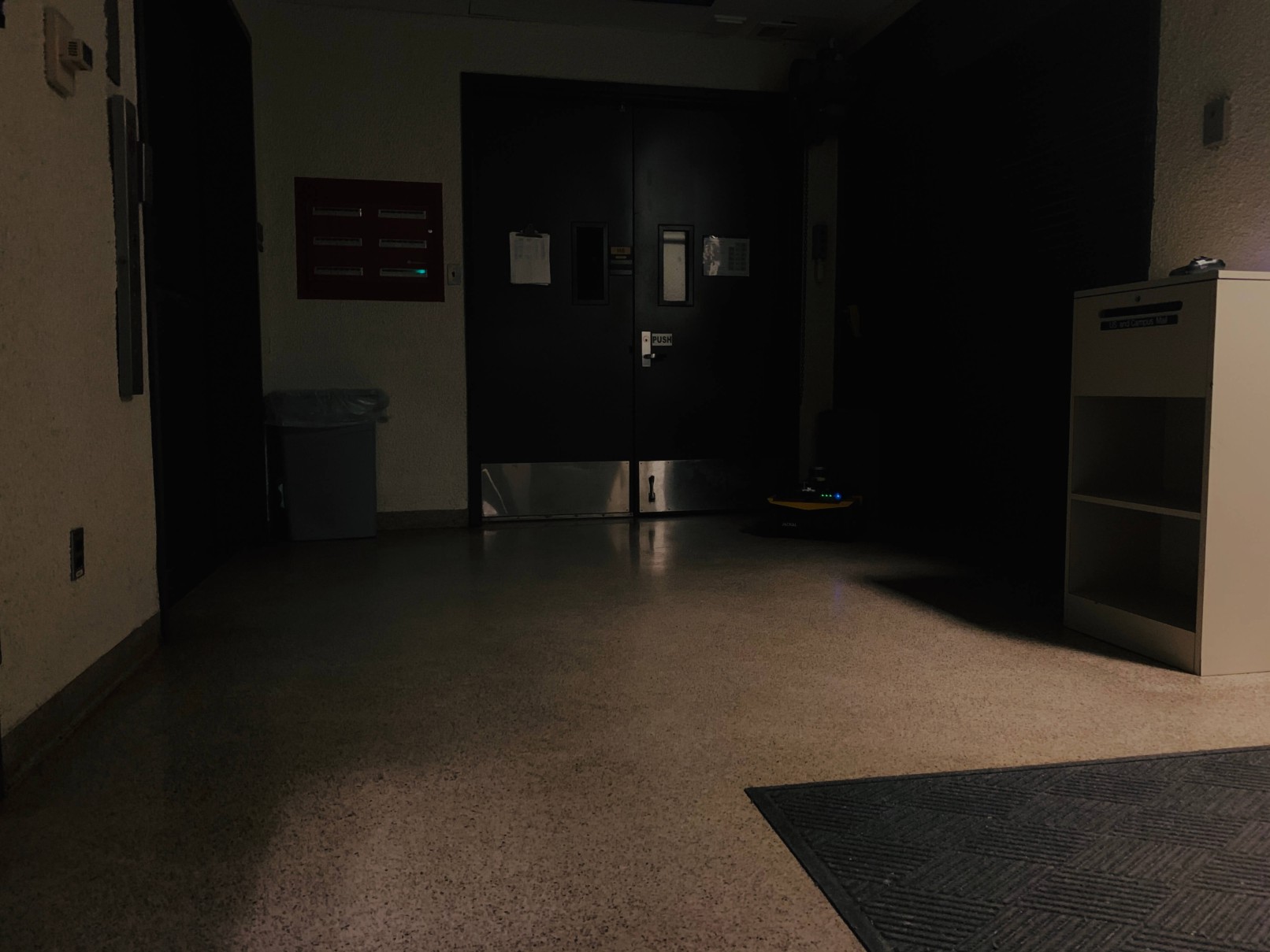}
        \caption{}
        \label{fig:loc_b}
    \end{subfigure}\hfill
    \begin{subfigure}{0.195\linewidth} 
        \includegraphics[width=\linewidth]{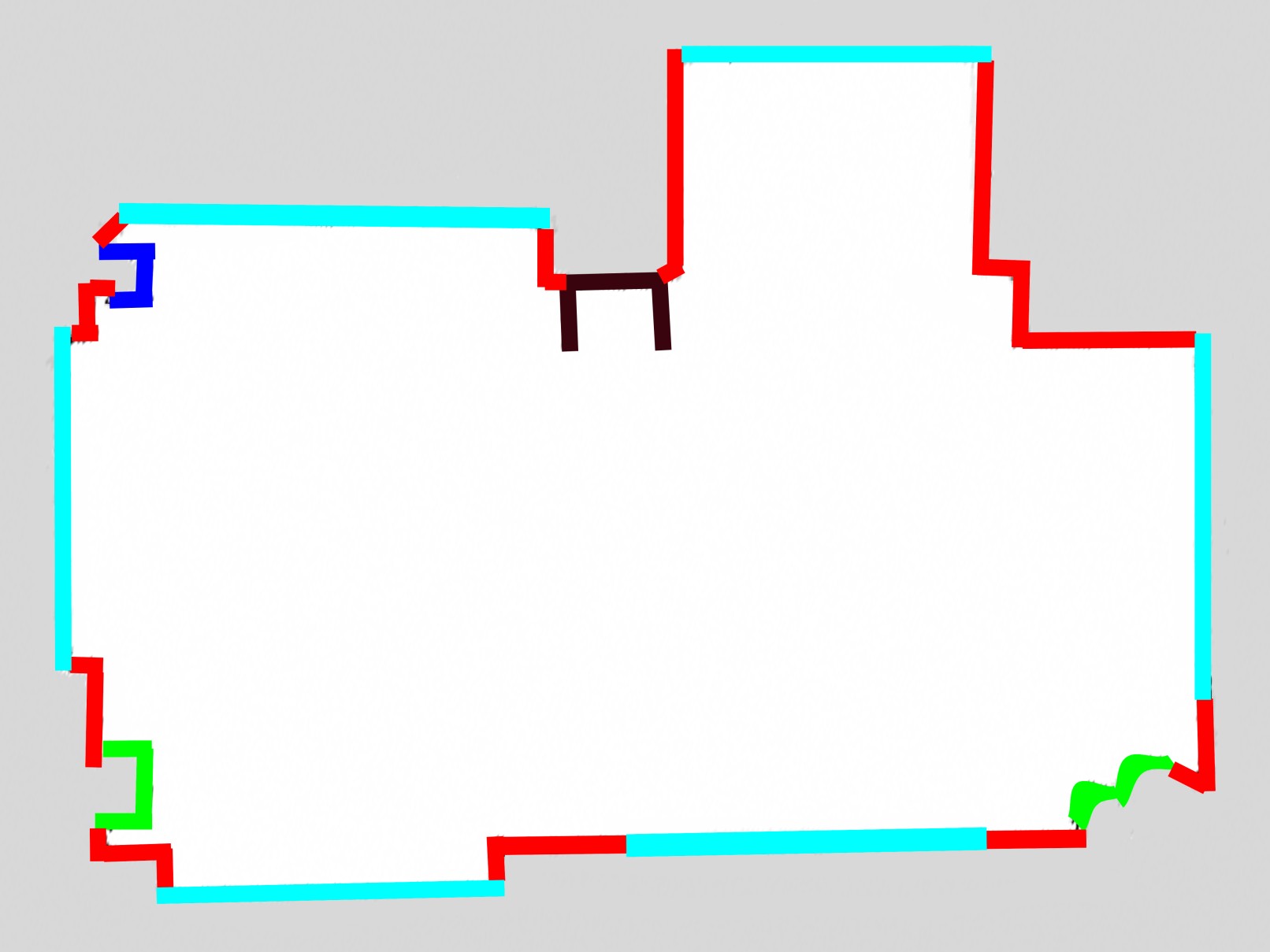}
        \caption{}
        \label{fig:loc_b_gt}
    \end{subfigure}\hfill
    \begin{subfigure}{0.195\linewidth} 
        \includegraphics[width=\linewidth]{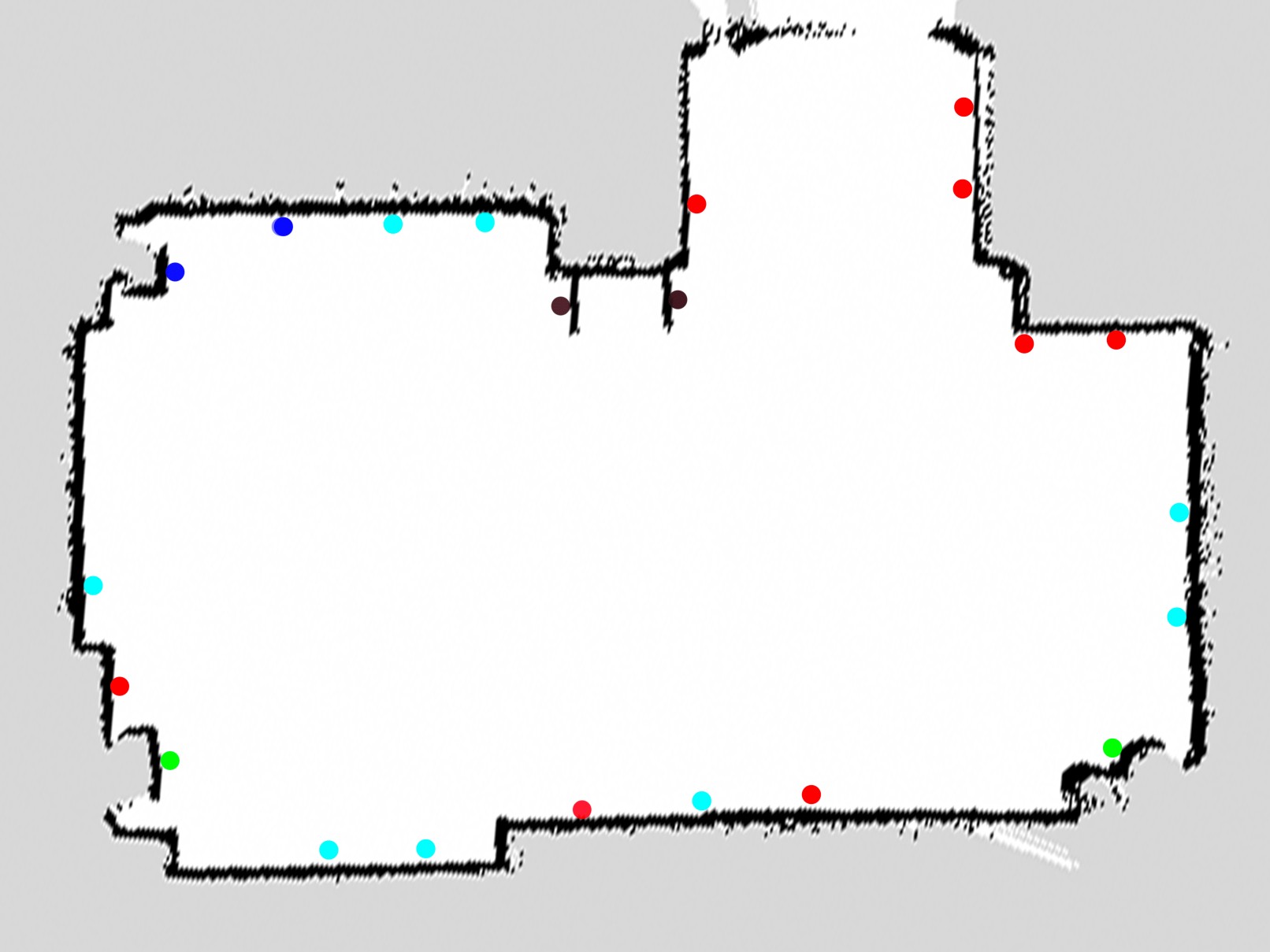}
        \caption{}
        \label{fig:loc_b_1}
    \end{subfigure}\hfill
    \begin{subfigure}{0.195\linewidth} 
        \includegraphics[width=\linewidth]{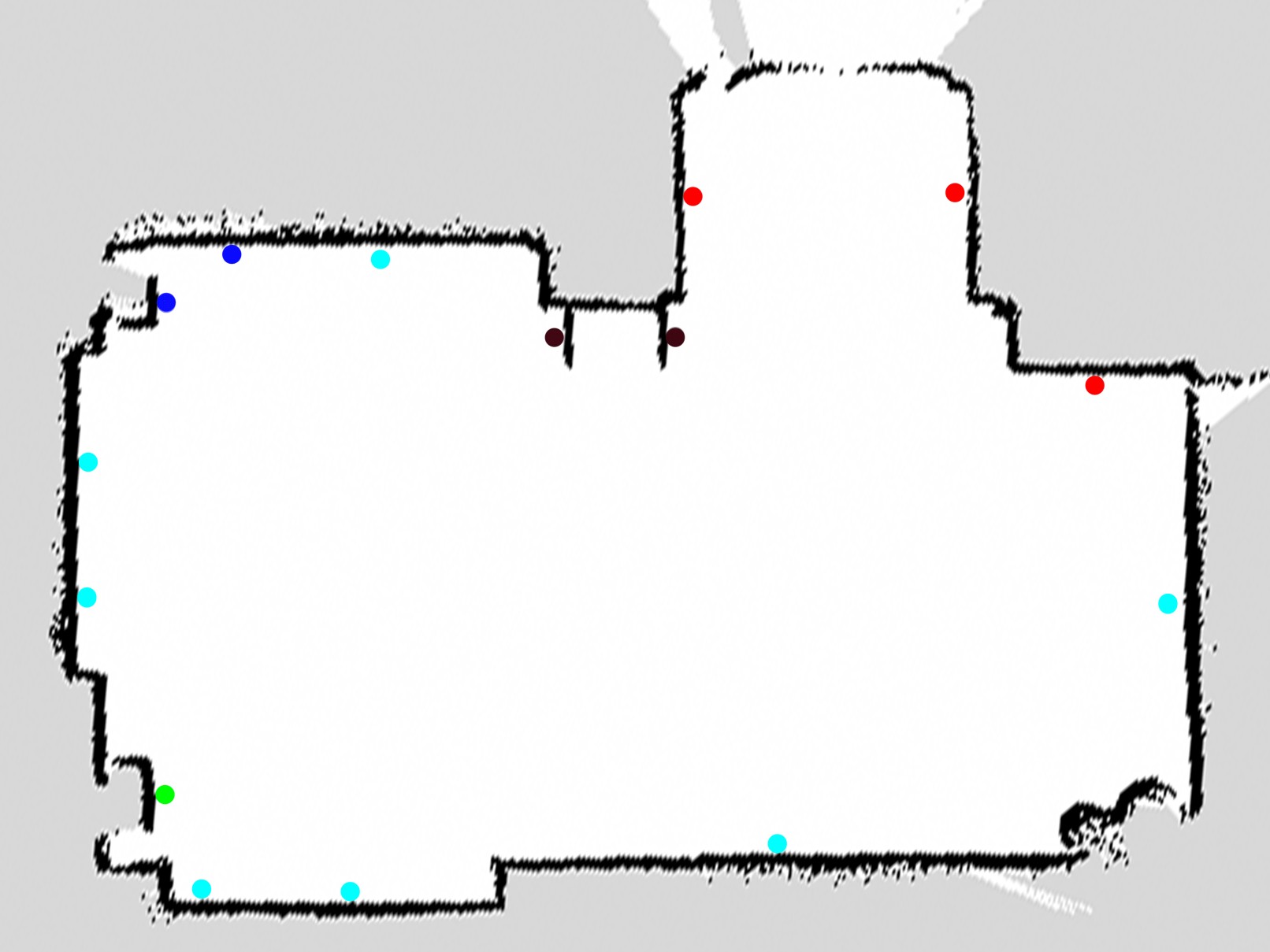}
        \caption{}
        \label{fig:loc_b_2}
    \end{subfigure}\hfill
    \begin{subfigure}{0.195\linewidth} 
        \includegraphics[width=\linewidth]{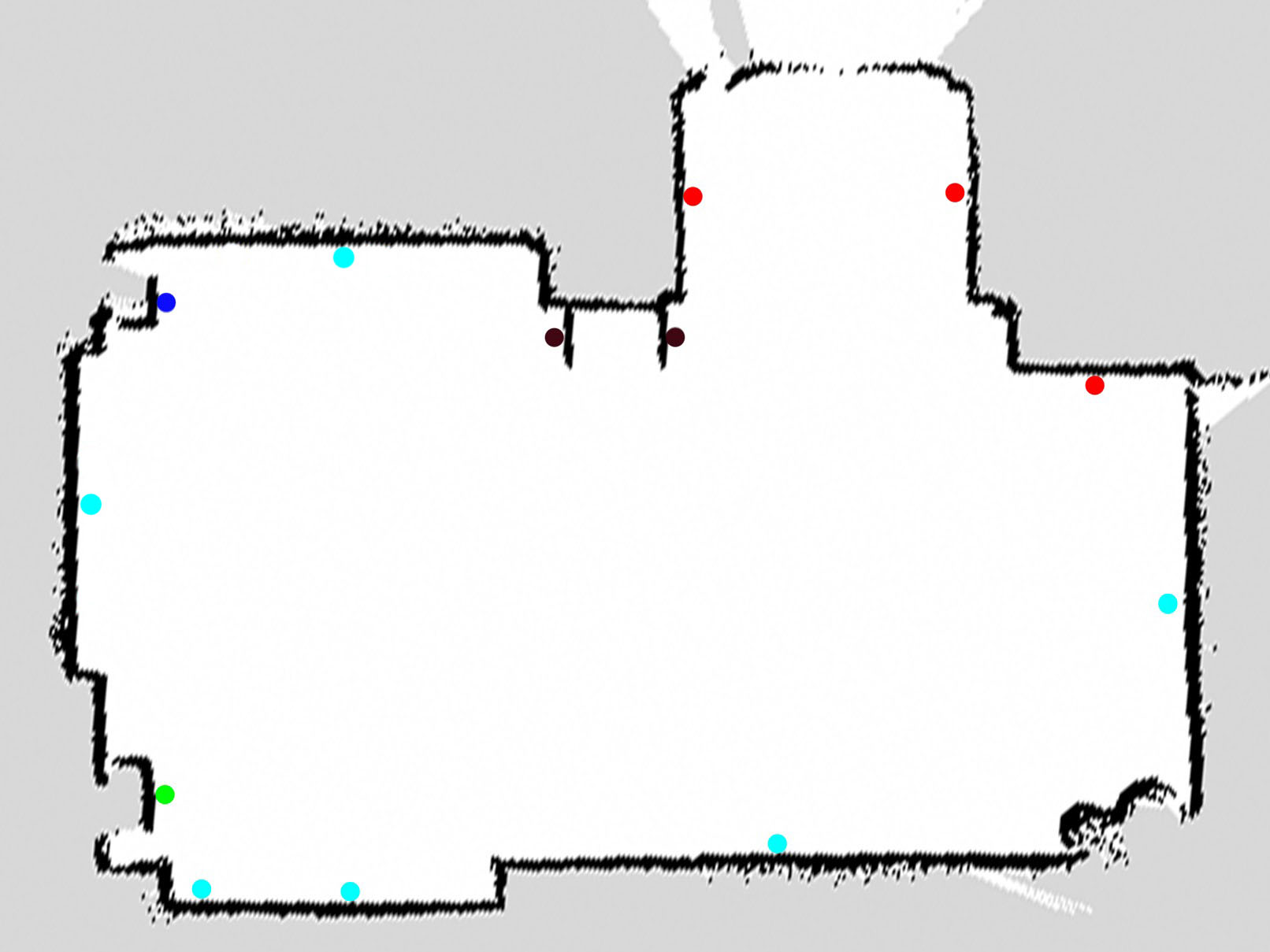}
        \caption{}
        \label{fig:loc_b_3}
    \end{subfigure}\hfill
\caption{Material maps constructed in field experiments in an unknown environment: (a) picture of the hallway environment used for the first experiment, (b) color-coded ground truth information of the various materials in the first environment; the material maps constructed with distance parameters as (c) 0.5 $m$, (d) 1 $m$, and (e) 2 $m$; (f) picture of the gateway environment used for the second experiment, (g) color-coded ground truth information of the various materials in the second environment; the material maps constructed with distance parameters as (h) 0.5 $m$, (i) 1 $m$, and (j) 2 $m$. The color coding for material maps and ground truth are as follows: \textcolor{SkyBlue}{metal (sky blue)}, \textcolor{Green}{plastic (green)}, \textcolor{Red}{concrete (red)}, \textcolor{purple}{glass (purple)}, \textcolor{blue}{cardboard (blue)}, \textcolor{yellow}{wall (yellow)}, and \textcolor{Sepia}{wood (brown)}. (The readers are requested to refer to the supplementary video for demonstration of the robot exploration. YouTube link: \url{https://youtu.be/4VGntBByJWE})
}
\label{fig:experiment_result}    
\end{figure*}

\section{Experiment and Results}
\label{sec:results}
This section shows the experimental results of the proposed system in real-world environments. The robot was deployed into desolated and cluttered environments with poor lighting conditions, and the material map of the environment was constructed. Multiple trials were conducted in the same environment with different $\gamma$ values.

\subsection{Experiment in an indoor hallway}
The first set of experiments was conducted in an isolated hallway inside a building. The environment used for the experiment is shown in Fig. \ref{fig:location_a}. This environment was preferred due to the presence of multiple doors and other entry points. A couple of doors were in line with the wall. In addition to the doors, the hallway had various objects like cylinders, cardboard boxes, cabinets (metal), wooden frames and so on. The ground truth information about the various materials in this region is shown in Fig. \ref{fig:loc_a_gt} using different colors for different materials. 

During the experiment, three different values of $\gamma$: 0.5, 1, and 2 $m$ were used. The corresponding material maps constructed are shown in Fig. \ref{fig:loc_a_1}, \ref{fig:loc_a_2} and \ref{fig:loc_a_3}. From the material map in  Fig. \ref{fig:loc_a_1}, it can be seen that almost all the materials present have been identified. There were some misclassifications in the bottom concrete wall, where the concrete was identified as metal and cardboard. Due to the congestion in the environment, the robot was unable to reach the cardboard box (present right next to the metal cabinet) and hence the cardboard was never identified. Sometimes due to the errors in the localization, the robot was unable to orient itself in the right direction for tapping and the solenoid failed to tap on the material. This was identified as an empty tap and no information was added to the material map at that location. This issue was found to be common across other values of the distance parameter too. From Fig. \ref{fig:loc_a_2}, it can be seen that even with the distance parameter as $1~m$, all the materials have been captured so as using $0.5~m$ (except for the few points missed due to empty taps). When the distance parameter is further increased to $2~m$, some materials were missed. In Fig. \ref{fig:loc_a_3}, the door in the top boundary which was present at the same level as the wall was never tapped and it was not updated in the material map. The experiment was terminated halfway through the hallway in all the three trails once sufficient data to analyze was collected. 

\begin{table*}[h!]
\centering
\caption{Accuracy of the proposed material recognition system in the exploration experiments with different distance parameter $\gamma$ and the time taken (Exp. 1 - the indoor hallway; Exp. 2 - the gateway of building). The materials that were not present in an experiment are marked with `x'.}
  \begin{tabular}{|l| c c c c c c c | c |c|} 
     \hline
     Exp. \# & $\gamma$  &  Metal & Wood & Cardboard & Plastic & Concrete & Wall  &  Overall Accuracy & Time Taken (s) \\ 
     \hline 
     \hline
     1 &  		0.5 & 10/11 & 9/9 & x & x & 8/8 & 5/7  & 0.91  & 645\\ 
     \hline
     1 &  		1  & 5/6 & 6/6 & x & x & 7/7 & 4/5 & 0.91 & 432\\ 
     \hline
     1 &  		2  & 5/6 & 5/5 & x & x & 5/5 & 4/5  & 0.90 & 380\\  
     \hline
     2 &  		0.5  & 8/15 & 2/2 & 1/1 & 2/2 & 7/8 & x & 0.71 & 463\\ 
     \hline
     2 &  		1  & 7/10 & 2/2 & 1/1 & 1/2 & 3/3 & x  & 0.77 & 322\\ 
     \hline
     2 &  		2  & 6/6 & 2/2 & 1/1 & 1/2 & 3/3 & x & 0.92 & 284\\  
    \hline
    \hline
      Total &   -  & 41/54 & 26/26 & 3/3 & 4/6 & 33/34 & 13/17  & 0.86 & -\\   
     \hline 
     \hline
  \end{tabular}
\label{tab:result_Experiment} 
\end{table*}

\subsection{Experiment at the dark gateway of a building}
Another set of experiments was conducted in an unlighted gateway of a building where various objects like doors (metal), trash bins, cabinets (wood) are placed. The experiment environment is shown in Fig. \ref{fig:loc_b}. 

Similar to the previous experiment, three trials were performed with $\gamma$ as 0.5, 1 and 2. Fig. \ref{fig:loc_b_1} shows the material map built with $\gamma$ as 0.5. Like the previous experiment, all the materials have been captured through taps along with some misclassifications. But, when the value of $\gamma$ is increased to 1 or 2, it can be seen that some parts of the concrete wall on the bottom side are not found. This experiment was performed completely until the entire space was explored, i.e there were no new frontier boundaries.

The classification accuracy, time taken and the value of $\gamma$ from all the trials have been summarized in Table \ref{tab:result_Experiment} (experiment 1 - indoor hallway; experiment 2 - the gateway of a building). Accuracy gives the count of the number of taps that were classified correctly to the number of taps performed corresponding to each material (this includes the taps that failed due to orientation issues). The accuracy depends on the nature of the objects present and is not influenced by the value of $\gamma$.  It is interesting to note that as the value of $\gamma$ increases, the time taken reduces significantly. We also observed that with a smaller value of $\gamma$, all the materials presented could be identified. This substantiates the trade-off that exists between the time taken and capability to map all the materials, and indicates that it would be ideal to select the value based on the scenario and the time constraints. 

\section{Conclusion and Future Work}
\label{sec:conclusion}
In this paper, we have presented a tapping system for a mobile robot that can be used for mapping various materials such as wood, plastic, metal, glass, wall, etc. in an unknown environment. We have discussed the various components of the system including the hardware design, the sound classification method, and the robot control algorithm for autonomous navigation and mapping.

Through real experiments mimicking search and rescue scenarios, we have demonstrated that using our proposed tapping system, we can classify the materials with an accuracy of approximately 97.45\% in identifying materials of known objects and an average accuracy of 86.95\% in identifying materials in unknown environments. The obtained materials map integrated with the SLAM map of the robot is not only useful for improving the robot's autonomy but also useful for preplanning robot operations in various applications such as search and rescue.

Currently, the classification accuracy under unknown environments is limited by the size of the dataset and the variations in tapping sounds. As future work, we intend to extend the size of the dataset with more samples and variations like different materials in contact with one another. Another potential direction for future research would be is to use the objects detected as features for solving SLAM problems like loop closure. 





\bibliographystyle{IEEEtran}
\bibliography{references}

\end{document}